\documentclass[10pt,twocolumn,letterpaper]{article}

\usepackage{iccv}
\usepackage{times}
\usepackage{epsfig}
\usepackage{graphicx}
\usepackage{amsmath}
\usepackage{amssymb}
\usepackage{amsfonts}

\usepackage[pagebackref=true,breaklinks=true,letterpaper=true,colorlinks,bookmarks=false]{hyperref}
\iccvfinalcopy
\def\iccvPaperID{}

\usepackage{algorithmic}
\usepackage{algorithm}
\usepackage{array}
\usepackage{textcomp}
\usepackage{url}
\usepackage{verbatim}
\usepackage{booktabs}
\usepackage{multirow}
\usepackage{makecell}
\usepackage{pifont}
\usepackage{float}
\usepackage{enumitem}
\usepackage[table]{xcolor}
\usepackage{tabulary,overpic}
\usepackage{color}
\usepackage{dblfloatfix}
\usepackage{nicefrac}
\usepackage{relsize}

\newlength\savewidth

\newcolumntype{x}[1]{>{\centering\arraybackslash}p{#1pt}}
\newcolumntype{y}[1]{>{\raggedright\arraybackslash}p{#1pt}}
\newcolumntype{z}[1]{>{\raggedleft\arraybackslash}p{#1pt}}

\definecolor{baselinecolor}{gray}{.92}

\definecolor{demphcolor}{gray}{.2}

\definecolor{demphcolorinline}{gray}{.3}

\definecolor{demphcolor1}{gray}{.6}

\renewcommand{\paragraph}[1]{\vspace{1.25mm}\noindent\textbf{#1}}

\newcommand{\app}{\raise.17ex\hbox{$\scriptstyle\sim$}}

\newcommand{\authorskip}{\hspace{5mm}}

\usepackage[capitalize]{cleveref}
\crefname{section}{Sec.}{Secs.}
\Crefname{section}{Section}{Sections}
\Crefname{table}{Table}{Tables}
\crefname{table}{Table}{Tables}

\usepackage{caption}
\usepackage{subcaption}

\begin{document}

\title{Revisiting Shadow Detection from a Vision-Language Perspective}

\author{
	\begin{tabular}{c}
		Yonghui Wang\textsuperscript{1}
		\authorskip Shaokai Liu\textsuperscript{2}
		\authorskip Wengang Zhou\textsuperscript{3}
		\authorskip Hao Feng\textsuperscript{3}
		\authorskip Houqiang Li\textsuperscript{1} \\[2mm]
		{\fontsize{10.4pt}{9.84pt}\selectfont
		\textsuperscript{1} Department of Electronic Engineering and Information Science, University of Science and Technology of China} \\
		{\fontsize{10.4pt}{9.84pt}\selectfont
		\textsuperscript{2} School of Management, Hefei University of Technology} \\
		{\fontsize{10.4pt}{9.84pt}\selectfont
		\textsuperscript{3} School of Artificial Intelligence and Data Science, University of Science and Technology of China}
	\end{tabular}
}

\maketitle
\ificcvfinal\thispagestyle{empty}\fi

\begin{abstract}
Shadow detection is commonly formulated as a vision-driven dense prediction problem, where models rely primarily on pixel-wise visual supervision to distinguish shadows from non-shadow regions.
However, this formulation can become unreliable in visually ambiguous cases, where similar dark regions may correspond either to cast shadows or to intrinsically dark surfaces, making visual evidence alone insufficient for establishing a stable decision rule.
In this work, we revisit shadow detection from a vision--language perspective and argue that robust prediction benefits from an explicit semantic reference beyond visual cues alone.
We propose SVL, a Shadow Vision--Language framework that uses language as an explicit semantic reference to disambiguate shadows from visually similar dark regions.
SVL aligns global image representations with shadow-related text embeddings through scene-level shadow ratio regression, and transfers this semantic guidance to dense prediction via global-to-local coupling and local patch-level constraints.
Built on a frozen DINOv3 image encoder, SVL learns only lightweight projection and decoding modules, yielding a parameter-efficient design with less than $1\%$ trainable parameters.
Extensive experiments on multiple shadow detection benchmarks, including dedicated hard-case evaluations, suggest strong overall performance and improved robustness under visually ambiguous conditions.
Code is available at \url{https://github.com/harrytea/SVL}.
\end{abstract}



\section{Introduction}
Shadow detection is a long-standing problem in computer vision and plays a critical role in a wide range of applications~\cite{sanin2012shadow,tiwari2016survey}.
Accurate shadow detection helps reduce illumination-induced uncertainty and enhances the robustness of downstream tasks, including scene understanding~\cite{alshammari2019multi}, autonomous driving~\cite{yang2024shadow}, and visual recognition~\cite{guo2017efficient}.
Despite substantial progress, shadow detection remains challenging in complex real-world scenes, particularly when cast shadows become difficult to distinguish from intrinsically dark non-shadow regions under ambiguous visual conditions.

\begin{figure}[t]
	\centering
	\includegraphics[width=1.0\linewidth]{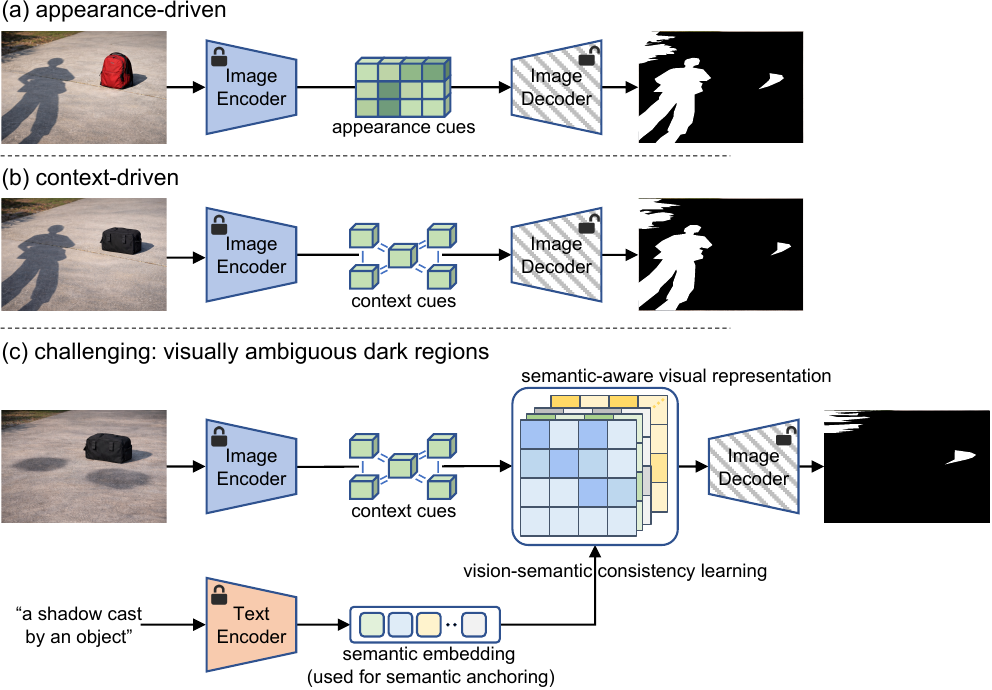}
    \caption{Progressive ambiguity in shadow detection.
    (a) Appearance-driven: shadows are separable from nearby objects using clear local appearance cues.
    (b) Context-driven: dark objects exhibit shadow-like appearance, making local cues unreliable and requiring broader contextual reasoning.
    (c) Challenging (visually ambiguous dark regions): shadows and intrinsically dark surfaces become highly similar under a single RGB observation, so appearance and context alone may not provide a stable decision criterion.}
	\label{fig:intro}
\end{figure}

Mainstream research commonly formulates shadow detection as a vision-centric dense prediction task.
Under this paradigm, deep learning--based methods~\cite{hosseinzadeh2018fast,wang2018stacked,zhu2018bidirectional,le2018a+} typically adopt encoder--decoder architectures to extract and fuse visual features for pixel-wise inference.
Based on how visual information is exploited, existing approaches fall into two categories: appearance-driven and context-driven.
The former relies on local cues, such as texture and color variations, whereas the latter leverages multi-scale or hierarchical representations to incorporate global spatial contexts.
Importantly, both categories are built upon the assumption that the input image contains sufficient visual evidence for reliable discrimination.
This assumption may fail in complex scenes: when visual evidence becomes ambiguous due to illumination, material properties, or occlusion, purely vision-based formulations tend to expose their inherent limitations, especially in visually ambiguous regions where shadows closely resemble non-shadow surfaces.

Figure~\ref{fig:intro} illustrates how the limitations of visually driven formulations manifest under increasing levels of visual ambiguity.
Figure~\ref{fig:intro}(a) shows an appearance-driven case, where shadows exhibit clear intensity attenuation and well-defined boundaries due to strong contrast with nearby objects (e.g., a red backpack).
In this scene, local appearance cues alone are sufficient for reliable shadow detection.
Figure~\ref{fig:intro}(b) depicts a context-driven case, where dark objects (e.g., a black backpack) mimic shadow appearance.
Here, local contrast is diminished, so models must rely more on context aggregation and long-range interactions to disambiguate shadows from objects.
Figure~\ref{fig:intro}(c) highlights a particularly challenging regime for purely visual reasoning.
In this highly ambiguous case, shadow regions and adjacent dark material surfaces exhibit strong visual similarity, making them difficult to distinguish even when contextual cues are considered.
Specifically, under a single RGB observation, image evidence from local appearance and global context may be insufficient to establish a stable decision rule for a subset of ambiguous cases.
In such cases, scaling up training data is a common recourse, yet it does not necessarily resolve the ambiguity in a conceptually grounded manner.
Instead, models may overfit dataset-specific correlations rather than learn a stable and transferable notion of ``shadow.''
These observations suggest that further progress may require not only stronger visual representations, but also an explicit semantic reference that specifies what should be regarded as ``shadow'' beyond visual cues.

Some recent studies have begun to move in this direction by introducing auxiliary cues, such as additional semantic supervision~\cite{zhou2024semantic} or prompt-based adaptation of foundation models~\cite{jie2025shadowadapter}.
These cues typically serve as implicit auxiliary signals within a vision-centric pipeline, rather than explicitly specifying the concept of ``shadow'' in a modality-independent manner.
In this work, we introduce language as an explicit semantic reference for shadow detection, with the goal of improving robustness in visually ambiguous cases.
We argue that shadow detection can benefit from going beyond purely visual cues and incorporating an explicit semantic specification of what constitutes ``shadow''.
Based on this insight, we propose \textbf{SVL}, a \textbf{S}hadow \textbf{V}ision--\textbf{L}anguage framework that enforces \emph{vision--semantic consistency learning} for dense shadow prediction.
Specifically, we instantiate this framework at both global and local granularities.

At the global level, we use shadow-related text embeddings as an explicit semantic reference to align the global visual representation.
This alignment is guided by a scene-level shadow ratio regression objective, which encourages the model to predict the overall extent of shadows at the image level, thereby infusing the global representation with semantic information and yielding a scene-level summary of shadow strength.
We then propagate this language-conditioned global representation to dense prediction through global-to-local coupling.
This coupling imposes a cross-granularity consistency constraint, encouraging patch-wise predictions to align with scene-level semantics rather than rely solely on local cues.

While global semantics provides a scene-level prior, it is often insufficient to resolve pixel-level confusions between cast shadows and intrinsically dark materials.
Therefore, at the local level, we introduce a local semantic constraint that regularizes pixel-wise features with text embeddings to enable fine-grained disambiguation.
This constraint acts as a patch-level semantic regularizer, reducing false positives on visually similar non-shadow regions.
The resulting multi-level consistency cues are then fed into a decoder for dense prediction.
To refine shadow boundaries, the decoder additionally incorporates shallow visual features.
This design preserves fine-grained boundary details while keeping the trainable modules compact, with less than $1\%$ of the total parameters updated during training.
Extensive experiments on several benchmarks show that SVL is particularly effective under visually ambiguous conditions, where cast shadows must be distinguished from visually similar dark non-shadow regions.

Our main contributions are summarized as follows:
\begin{itemize}
\item We present SVL, a new vision--language framework for shadow detection that enforces vision--semantic consistency through global semantic injection and local semantic constraint.
\item We introduce a scene-level shadow ratio regression objective that provides a global shadow prior for calibrating the language-guided global representation.
\item We provide a parameter-efficient implementation with less than $1\%$ trainable parameters, and extensive experiments show that it improves robustness under visually ambiguous conditions on multiple benchmarks.
\end{itemize}

\section{Related Work}
\subsection{Shadow Detection Methods}
Shadow detection has been extensively studied over the past decade, as accurate shadow detection is a crucial prerequisite for subsequent shadow removal~\cite{liu2023decoupled,niu2022boundary,chi2026cross,jung2009efficient}. 
Early methods mainly relied on physical image formation models and low-level appearance cues, such as dichromatic or bi-illuminant reflection models~\cite{maxwell2008bi}, spectral color ratios~\cite{huang2009moving}, and local intensity or texture statistics~\cite{zhu2010learning}. 
However, these methods are often limited by handcrafted feature design, leading to insufficient robustness in complex scenes.
With the development of deep learning, shadow detection has been commonly formulated as a vision-driven dense prediction task.
Existing methods typically exploit encoder--decoder architectures, multi-scale feature fusion, and context aggregation to improve pixel-wise prediction~\cite{wang2019moving,mohajerani2019shadow,hu2018direction,zhu2018bidirectional,jie2023rmlanet,zheng2019distraction}.
Recent studies further address ambiguity and supervision limitations through bias-aware modeling, dark-region enhancement, large-scale or synthetic data, and weakly supervised learning~\cite{zhu2021mitigating,zhu2022single,wu2025pay,qiaounder,hou2019large,hu2021revisiting,inoue2020learning,chen2020multi,yang2023silt}.
Transformer backbones and pretrained foundation models have also been explored to incorporate global context and high-level representations~\cite{wang2023detect,wang2024swinshadow,sun2025structure,zhou2024semantic,chen2023make,jie2025shadowadapter,qian2024omni,wang2024language}.

Despite these advances, semantic information is still inferred implicitly from visual patterns, and most methods remain vision-driven.
Consequently, performance often degrades in visually ambiguous cases, where dark objects, textured regions, or low-contrast non-shadow surfaces resemble cast shadows.
This exposes a key limitation: visual evidence alone may be insufficient to reliably disambiguate shadows from visually similar non-shadow regions.
Motivated by this limitation, we move beyond visual patterns by introducing language as an explicit semantic reference to specify what should be regarded as ``shadow''.

\subsection{Vision--Language Alignment}
Vision--language alignment has emerged as an effective paradigm for learning transferable visual representations by aligning images and natural language in a shared embedding space~\cite{radford2021learning,li2022blip}.
This paradigm enables strong semantic generalization and has been widely adopted in downstream recognition and dense prediction tasks, including image classification, object detection, segmentation, and referring image segmentation~\cite{gu2021open,mukhoti2023open,li2022grounded}. 
More recently, multimodal large language models further use language as a unified semantic interface by projecting visual features into language embedding spaces, enabling instruction-following, language-guided understanding, and reasoning over complex visual content~\cite{liu2023visual}.

Motivated by these advances, we introduce language into shadow detection as an explicit semantic reference to better handle cases where visual evidence is ambiguous.
Unlike physical objects, shadows arise from illumination effects and do not exhibit stable intrinsic shapes, leading to appearance variations across scenes.
As a result, relying on visual cues alone may be insufficient when shadows closely resemble dark, non-shadow surfaces.
These observations motivate using language as an explicit semantic reference to complement vision-driven modeling in shadow detection.

\subsection{Semantic Consistency Learning}
Semantic consistency learning is an effective paradigm for improving generalization by aligning latent features or model predictions of the same data under diverse viewing conditions~\cite{tarvainen2017mean,lin2022dual}.
These approaches encourage the model to produce similar features or predictions for each sample under modest perturbations, thereby suppressing unstable variations and emphasizing reliable cues.
Semantic consistency has been explored in a wide range of settings, including supervised and semi-supervised learning, as well as cross-modal representation learning~\cite{zhou2023semantically,hu2023cross}.
Early work primarily focused on classification tasks by enforcing prediction consistency across augmentations or complementary branches~\cite{tarvainen2017mean,chen2021semi}.
This paradigm was later extended to dense prediction (e.g., detection and segmentation) by imposing pixel-level spatial or contextual consistency, which often improves structural coherence and boundary quality~\cite{xu2022semi}.
Relatedly, vision--language learning promotes cross-modal agreement between images and text in a shared embedding space, enabling effective semantic transfer across modalities~\cite{li2022blip,alayrac2022flamingo}.

Existing consistency-based methods mainly focus on invariance to observational perturbations, often lacking an explicit semantic reference for resolving conceptual ambiguity.
Shadow detection highlights this limitation: shadows and visually similar non-shadow regions (e.g., dark objects) can be highly similar at the pixel level, rendering purely visual consistency insufficient for reliable disambiguation.
To bridge this gap, we propose vision--semantic consistency learning.
By aligning visual representations with shadow-related text embeddings, we use language as an explicit semantic reference to regularize dense shadow prediction in ambiguous cases.

\section{Method}

\begin{figure*}[t]
	\centering
	\includegraphics[width=1.0\linewidth]{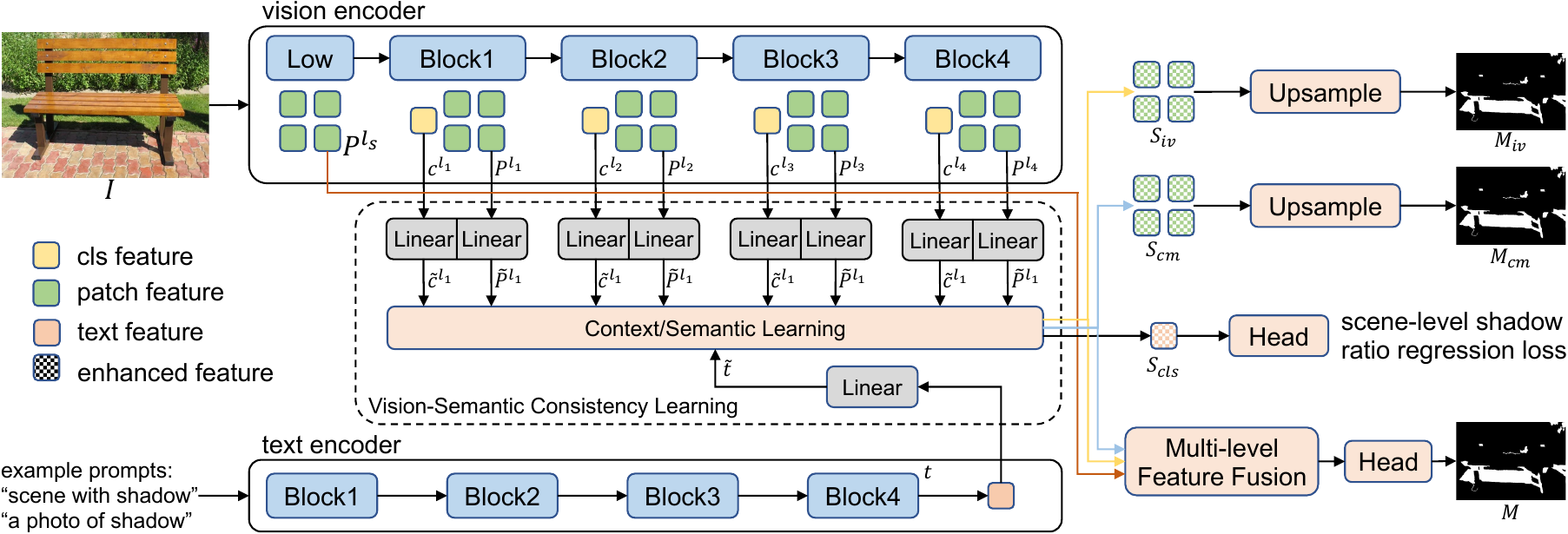}
    \caption{Overall architecture of SVL. A frozen vision encoder extracts multi-level \texttt{[CLS]} and patch tokens, while a frozen text encoder produces shadow-related text embeddings. SVL computes global and patch-wise consistency cues to inject language-guided information into dense prediction, applies auxiliary and scene-level supervision, and uses a lightweight fusion decoder to output the final shadow mask.}
    \label{fig:arch}
\end{figure*}

\subsection{Overview and Problem Formulation}
Given an input image $\mathbf{I}\in\mathbb{R}^{H\times W\times 3}$, shadow detection aims to predict a binary mask $\mathbf{M}\in\{0,1\}^{H\times W}$ indicating whether each pixel belongs to a shadow region.
Despite extensive research, shadow detection remains challenging in unconstrained scenes, where cast shadows often share similar appearances with intrinsically dark or textured non-shadow regions.
Most existing approaches are vision-driven, leveraging appearance cues together with contextual modeling.
While these cues are informative, low intensity alone is an ambiguous indicator of shadow, especially in the presence of dark objects or textured surfaces.
In such cases, incorporating broader context can help, yet ambiguity may persist under a single RGB observation.
To address this limitation, we reformulate shadow detection as a \emph{vision--semantic consistency learning} problem and propose SVL, a Shadow Vision--Language framework.
Our key insight is that shadows are illumination-induced phenomena rather than intrinsic surface properties.
Therefore, reliable detection should be guided not only by visual cues but also by an explicit semantic reference that specifies what constitutes a ``shadow''.
In this work, we use shadow-related text embeddings as an explicit semantic reference and enforce consistency between visual representations and text embeddings, while preserving contextual modeling in the visual encoder.

Figure~\ref{fig:arch} illustrates the overall pipeline of our framework.
Given an input image, a vision encoder extracts multi-level representations that encode shadow-related cues from different layers.
In parallel, a text encoder maps shadow-related prompts to text embeddings.
SVL enforces vision--semantic consistency between visual features and text embeddings via \emph{global semantic injection} and a \emph{local semantic constraint}.
To facilitate optimization, we supervise intermediate predictions with auxiliary losses.
We further introduce a scene-level shadow ratio regression objective that estimates image-level shadow strength, providing a global prior for separating cast shadows from visually similar non-shadow regions.
Finally, a multi-level feature aggregation module fuses language-guided global representations from deeper layers with shallow local details to produce the final mask.
The framework is trained jointly, yielding predictions that better align with the explicit semantic reference provided by text embeddings.
We describe each component and the corresponding training objectives next.

\subsection{Multi-level Visual Representations for Shadow Modeling}
\label{method_mvrfsm}
Shadow detection benefits from exploiting multi-level features extracted from different layers of a vision backbone.
Specifically, shallow layers capture fine boundary details, whereas deeper layers encode global illumination cues and broader scene context.
This observation motivates aggregating complementary cues across layers for robust shadow modeling.
We adopt a multi-level feature extraction scheme based on a transformer vision backbone.
Given an input image, the backbone produces a set of layer-wise representations $\{\mathbf{F}^l\}_{l=1}^{L}$, where the selected layers share a consistent token grid (i.e., the same spatial resolution).
Each representation $\mathbf{F}^l$ consists of a global \texttt{[CLS]} token and a set of patch tokens, and can be written as
\begin{equation}
\mathbf{F}^l = ( \mathbf{c}^l, \mathbf{P}^l ),
\end{equation}
where the global token $\mathbf{c}^l \in \mathbb{R}^{D}$ summarizes image-level context, and the patch tokens
\(
\mathbf{P}^l = \{\mathbf{p}_i^l\}_{i=1}^{N}, \ \mathbf{p}_i^l \in \mathbb{R}^{D}
\)
preserve spatially localized information for the $N$ image patches.
These two token types provide complementary cues for shadow modeling.
Specifically, \texttt{[CLS]} tokens $\mathbf{c}^l$ summarize global context, whereas patch tokens $\mathbf{P}^l$ retain spatial evidence for shadow localization.
Together, these multi-level visual representations serve as the visual inputs to the subsequent vision--semantic consistency modules.
In parallel, a text encoder maps shadow-related prompts to a text embedding $\mathbf{t} \in \mathbb{R}^{D}$, which is used as an explicit semantic reference in the subsequent modules.
We select four layers at different depths of the backbone to obtain the multi-level visual features
\(
\{\mathbf{c}^{l_i}, \mathbf{P}^{l_i}\}_{i=1}^{4}.
\)
In addition, to preserve boundary details in the final prediction, we incorporate an extra shallow-layer patch feature $\mathbf{P}^{l_s} \in \mathbb{R}^{N \times D}$ in the mask fusion stage.
This integration is implemented via a skip connection.

\begin{figure}[t]
	\centering
	\includegraphics[width=1.0\linewidth]{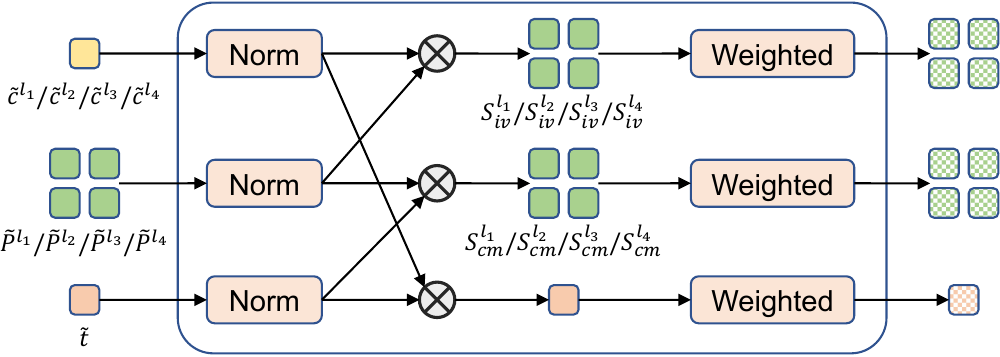}
    \caption{Overview of the proposed vision--semantic consistency learning. We compute multi-level consistency scores between shadow-related text embeddings, \texttt{[CLS]} tokens, and patch tokens, and fuse them into dense cues for shadow prediction.}
    \label{fig:consistency}
\end{figure}

\subsection{Vision-Semantic Consistency Learning}
Building upon the multi-level visual representations, we introduce a vision--semantic consistency learning framework that enforces consistency across modalities and granularities (Figure~\ref{fig:consistency}).
Specifically, it comprises two complementary pathways: (i) \textit{global semantic injection}, which injects scene-level semantic guidance into the global representation to guide dense prediction; and (ii) \textit{local semantic constraint}, which directly constrains pixel-wise features with text embeddings.
During training, shadow predictions are jointly regularized by the injected global guidance and the explicit semantic reference provided by the text embeddings.

\subsubsection{Global Semantic Injection}
Local visual cues alone can be insufficient to distinguish shadows from visually similar non-shadow regions.
As shown in Figure~\ref{fig:consistency}, we calibrate the global \texttt{[CLS]} token with the text embedding and then transfer the resulting global guidance to patch tokens.
A text encoder maps a shadow-related prompt to a text embedding $\mathbf{t} \in \mathbb{R}^{D}$.
We project the text embedding into the visual feature space via a linear transformation:
\begin{equation}
\tilde{\mathbf{t}} = \mathbf{W}_t \mathbf{t},
\end{equation}
where $\mathbf{W}_t \in \mathbb{R}^{D \times D}$ is a learnable projection matrix.
We apply $\ell_2$ normalization to obtain $\hat{\mathbf{t}} \in \mathbb{R}^{D}$.
For each selected layer $l_i$, we project the global token $\mathbf{c}^{l_i}$ and patch tokens $\mathbf{P}^{l_i}$ into the same feature space:
\begin{equation}
\tilde{\mathbf{c}}^{l_i} = \mathbf{W}_c \mathbf{c}^{l_i},
\qquad
\tilde{\mathbf{p}}_j^{l_i} = \mathbf{W}_p \mathbf{p}_j^{l_i},
\end{equation}
where $\mathbf{W}_c, \mathbf{W}_p \in \mathbb{R}^{D \times D}$ are learnable projection matrices.
We apply $\ell_2$ normalization to obtain $\hat{\mathbf{c}}^{l_i}$ and $\hat{\mathbf{p}}_j^{l_i}$.
We compute a \texttt{[CLS]}--text consistency score to inject semantic information from the text embedding into the global token:
\begin{equation}
s_{\mathrm{cls}}^{l_i} = \beta \, \langle \hat{\mathbf{c}}^{l_i}, \hat{\mathbf{t}} \rangle.
\end{equation}
Here, $s_{\mathrm{cls}}^{l_i} \in \mathbb{R}$ is the scalar consistency score and $\beta$ is a learnable scaling factor.
This score provides an explicit cross-modal semantic signal, and its supervision by the scene-level shadow ratio regression objective encourages the global token $\hat{\mathbf{c}}^{l_i}$ to encode shadow-related semantics consistent with the text reference.
We then transfer this semantically calibrated global information to patch tokens by computing patch--\texttt{[CLS]} consistency:
\begin{equation}
s^{l_i}_{j,\mathrm{gi}} = \alpha \, \langle \hat{\mathbf{p}}_j^{l_i}, \hat{\mathbf{c}}^{l_i} \rangle,
\end{equation}
where $s^{l_i}_{j,\mathrm{gi}} \in \mathbb{R}$ denotes the patch-wise consistency score and $\alpha$ is a learnable scaling factor.
This propagation transfers globally injected semantic information to patch-wise cues, providing global guidance for dense prediction in visually ambiguous regions.

\subsubsection{Local Semantic Constraint}
While global propagation provides scene-level guidance, it may miss fine boundary details and other localized variations.
As shown in Figure~\ref{fig:consistency}, we therefore impose a direct patch--text consistency constraint to calibrate local evidence with the text embedding.
Concretely, we compute a patch--text consistency score to directly regularize patch features with the text embedding.
Using the same normalized text embedding $\hat{\mathbf{t}} \in \mathbb{R}^{D}$, we measure its similarity to each patch feature:
\begin{equation}
s^{l_i}_{j,\mathrm{lc}} = \gamma \, \langle \hat{\mathbf{p}}_j^{l_i}, \hat{\mathbf{t}} \rangle,
\end{equation}
where $s^{l_i}_{j,\mathrm{lc}} \in \mathbb{R}$ denotes the cross-modal consistency score and $\gamma$ is a learnable scaling factor.
Unlike global propagation, which relies on the \texttt{[CLS]} token as an intermediary, this term directly regularizes each patch token against the text embedding.
Intuitively, true shadow regions should yield higher patch--text similarity, whereas visually similar dark non-shadow regions should yield lower similarity.
As a result, this term acts as a fine-grained regularizer that suppresses shadow-like activations on non-shadow regions.
We combine the propagated patch--\texttt{[CLS]} cues ($s^{l_i}_{j,\mathrm{gi}}$) with the direct patch--text cues ($s^{l_i}_{j,\mathrm{lc}}$) to form complementary consistency cues for dense prediction.

\subsection{Multi-level Feature Fusion}
Building upon vision--semantic consistency learning, we design a multi-level fusion module that integrates patch-wise consistency maps from multiple layers for predicting the final shadow mask.
For each selected layer $l_i$, we obtain two patch-wise consistency maps from global semantic injection (patch--\texttt{[CLS]}) and local semantic constraint (patch--text).
We denote these patch-level consistency maps as
\(
\mathbf{S}_{\mathrm{gi}}^{l_i} = \{ s^{l_i}_{j,\mathrm{gi}} \}_{j=1}^{N}
\)
and
\(
\mathbf{S}_{\mathrm{lc}}^{l_i} = \{ s^{l_i}_{j,\mathrm{lc}} \}_{j=1}^{N},
\)
where $\mathbf{S}_{\mathrm{gi}}^{l_i}, \mathbf{S}_{\mathrm{lc}}^{l_i} \in \mathbb{R}^{N}$, and each entry corresponds to one of the $N$ patch locations.
To aggregate information across layers, we compute the aggregated consistency maps as weighted sums with learnable fusion weights:
\begin{equation}
\mathbf{S}_{\mathrm{gi}} = \sum_{i=1}^{K} \omega_i^{\mathrm{gi}} \, \mathbf{S}_{\mathrm{gi}}^{l_i},
\qquad
\mathbf{S}_{\mathrm{lc}} = \sum_{i=1}^{K} \omega_i^{\mathrm{lc}} \, \mathbf{S}_{\mathrm{lc}}^{l_i},
\end{equation}
where $K$ denotes the number of selected layers and $\omega_i^{\mathrm{gi}}, \omega_i^{\mathrm{lc}} \in \mathbb{R}$ are learnable scalar weights that control the contribution of each layer.
To preserve boundary details, we additionally incorporate a shallow patch feature $\mathbf{P}^{l_s} \in \mathbb{R}^{N \times D}$ via a skip connection.
We concatenate $\mathbf{S}_{\mathrm{gi}}$, $\mathbf{S}_{\mathrm{lc}}$, and $\mathbf{P}^{l_s}$ and pass them to a refinement head to predict the final mask.
This design jointly leverages global guidance and local boundary details for reliable dense prediction.

\subsection{Training Objective}
The proposed framework is trained end-to-end with auxiliary dense supervision and a scene-level shadow ratio regression objective.
Given an input image, the network outputs two auxiliary logit maps, $\mathbf{M}_{\mathrm{gi}}$ and $\mathbf{M}_{\mathrm{lc}}$, and a final fused logit map $\mathbf{M}$.
In addition to dense predictions, the network predicts a global scalar $s$ for regressing the image-level shadow ratio.

\subsubsection{Class-Balanced Weighted BCE Loss}
Following prior work~\cite{zheng2019distraction,chen2020multi,wang2024swinshadow}, we apply a class-balanced weighted binary cross-entropy loss $\mathcal{L}_{\mathrm{wcp}}(\cdot,\cdot)$ to both auxiliary maps and the final fused prediction.
Let $\mathbf{Y}\in\{0,1\}^{H\times W}$ be the ground-truth mask, and let $\mathbf{Z}\in[0,1]^{H\times W}$ be the predicted probability map obtained after sigmoid activation.
We define $\omega$ as the non-shadow-to-shadow pixel ratio and apply a global scaling factor $\beta_{\mathrm{back}}$.
The loss is then formulated as
\begin{align}
\mathcal{L}_{\mathrm{wcp}}
=
-\beta_{\mathrm{back}}
\Big(
\omega\,\mathbf{Y}\log(\mathbf{Z}+\varepsilon)
+
(1-\mathbf{Y})\log(1-\mathbf{Z}+\varepsilon)
\Big),
\end{align}
where the loss is computed element-wise and averaged over spatial locations, and $\varepsilon$ is a numerical stabilizer.

\subsubsection{Scene-level Shadow Ratio Regression Loss}
We introduce a scene-level regularizer that regresses the shadow ratio of the input image.
Let $r\in[0,1]$ denote the ground-truth shadow ratio computed from the mask $\mathbf{Y}$.
The network predicts a scalar $s$, from which we obtain $\hat{r}=\sigma(s)\in[0,1]$.
We use the following weighted Smooth-$\ell_{1}$ loss:
\begin{equation}
\mathcal{L}_{\mathrm{ratio}}
=
(r+\delta)^{\kappa}\cdot \mathrm{Smooth}\text{-}\ell_{1}(\hat{r}-r),
\end{equation}
which assigns higher weights to images with larger shadow ratios.
We set $\delta=0.05$ and $\kappa=0.5$ in all experiments.

\subsubsection{Overall Objective}
The overall objective is defined as
\begin{equation}
\mathcal{L}
=
\mathcal{L}_{\mathrm{final}}
+
\lambda_{\mathrm{gi}}(t)\,\mathcal{L}_{\mathrm{gi}}
+
\lambda_{\mathrm{lc}}(t)\,\mathcal{L}_{\mathrm{lc}}
+
\lambda_{\mathrm{ratio}}\,\mathcal{L}_{\mathrm{ratio}}.
\end{equation}
Here $\mathcal{L}_{\mathrm{final}}$ denotes the weighted BCE loss on the final prediction, while $\mathcal{L}_{\mathrm{gi}}$ and $\mathcal{L}_{\mathrm{lc}}$ denote the corresponding auxiliary losses for the two intermediate maps. 
The weights of these two auxiliary losses are linearly decayed from $1$ to $0$ throughout training.
The coefficient $\lambda_{\mathrm{ratio}}$ weights the ratio regression loss, and we set $\lambda_{\mathrm{ratio}}=0.25$ in all experiments.
All terms are optimized jointly in an end-to-end manner.

\begin{table*}[t]
\centering
\renewcommand\arraystretch{1.1}
\setlength{\tabcolsep}{8pt}
\caption{Quantitative comparison with representative shadow detection methods on three benchmarks.
``BER'' denotes the balanced error rate.
``S'' and ``NS'' denote the BER values on shadow and non-shadow regions, respectively.
Best and second-best results are highlighted in \textbf{bold} and \underline{underlined}, respectively.
``$\downarrow$'' indicates lower is better.}
\label{tab:quantitative}

\scalebox{0.85}{
\begin{tabular}{l c ccc ccc ccc}
\toprule
\multirow{2}{*}{\textbf{Method}} & \multirow{2}{*}{\textbf{Year}} 
& \multicolumn{3}{c}{\textbf{SBU}} 
& \multicolumn{3}{c}{\textbf{UCF}} 
& \multicolumn{3}{c}{\textbf{ISTD}} \\
\cmidrule(lr){3-5} \cmidrule(lr){6-8} \cmidrule(lr){9-11}
& 
& BER$\downarrow$ & S$\downarrow$ & NS$\downarrow$
& BER$\downarrow$ & S$\downarrow$ & NS$\downarrow$
& BER$\downarrow$ & S$\downarrow$ & NS$\downarrow$ \\
\midrule
Unary-Pairwise~\cite{guo2011single} & 2011 & 25.03 & 36.26 & 13.80 & - & - & - & - & - & - \\
scGAN~\cite{wang2018stacked}       & 2018 & 9.10 & 8.39 & 9.69 & 11.50 & 7.74 & 15.30 & 4.70 & 3.22 & 6.18 \\
patched-CNN~\cite{hosseinzadeh2018fast} & 2018 & 11.56 & 15.60 & 7.52 & - & - & - & - & - & - \\
ST-CGAN~\cite{wang2018stacked}     & 2018 & 8.14 & 3.75 & 12.53 & 11.23 & \underline{4.94} & 17.52 & 3.85 & 2.14 & 5.55 \\
BDRAR~\cite{zhu2018bidirectional}       & 2018 & 3.64 & 3.40 & 3.89 & 7.81 & 9.69 & 5.44 & 2.69 & \textbf{0.50} & 4.87 \\
DSC~\cite{hu2018direction}         & 2018 & 5.59 & 9.76 & \textbf{1.42} & 10.38 & 11.72 & \textbf{3.04} & 3.42 & 3.85 & 3.00 \\
ADNet~\cite{le2018a+}       & 2018 & 5.37 & 4.45 & 6.30 & 9.25 & 8.37 & 10.14 & - & - & - \\
DC-DSPF~\cite{wang2018densely}     & 2019 & 4.90 & 4.70 & 5.10 & 7.90 & 6.50 & 9.30 & - & - & - \\
DSD~\cite{zheng2019distraction}         & 2019 & 3.45 & 3.33 & 3.58 & 7.52 & 9.52 & 5.53 & 2.17 & 1.36 & 2.98 \\
MTMT~\cite{chen2020multi}        & 2020 & 3.15 & 3.73 & 2.57 & 7.47 & 10.31 & \underline{4.63} & 1.75 & 1.36 & 2.14 \\
FDRNet~\cite{zhu2021mitigating}      & 2021 & 3.04 & 2.91 & 3.18 & 7.28 & 8.31 & 6.26 & 1.55 & 1.22 & 1.88 \\
SDCM~\cite{zhu2022single}        & 2022 & 2.94 & 3.54 & 2.35 & 6.73 & 8.15 & 5.31 & 1.44 & 1.19 & 1.69 \\
R2D~\cite{valanarasu2023fine}         & 2023 & 3.15 & 2.74 & 3.56 & 6.96 & 8.32 & 5.60 & 1.69 & \underline{0.59} & 2.79 \\
RMLANet~\cite{jie2023rmlanet}     & 2023 & 2.97 & 2.53 & 3.42 & 6.41 & 6.69 & 6.14 & 1.01 & 0.68 & 1.34 \\
SwinShadow~\cite{wang2024swinshadow}  & 2024 & 2.78 & 2.78 & 2.79 & \underline{5.99} & 6.57 & 5.40 & 1.22 & 1.05 & 1.40 \\
SaT~\cite{zhou2024semantic}   & 2024 & 3.02 & 3.68 & 2.36 & 7.18 & 9.14 & 5.22 & 1.53 & 1.24 & 1.82 \\
UTS~\cite{qiaounder}           & 2025 & - & - & - & - & - & - & 1.32 & 0.96 & 1.67 \\
STNet~\cite{sun2025structure}   & 2025 & \underline{2.74} & 3.16 & \underline{2.32} & 6.36 & 7.58 & 5.14 & 1.15 & 1.03 & 1.27 \\
ShadowAdapter~\cite{jie2025shadowadapter}   & 2025 & 2.80 & \underline{2.50} & 3.10 & 6.24 & 5.67 & 6.82 & \textbf{0.86} & \underline{0.59} & \underline{1.13} \\
\midrule
\textbf{Ours} & - & \textbf{2.43} & \textbf{2.24} & 2.62 & \textbf{5.69} & \textbf{4.72} & 6.66 & \underline{0.91} & 0.84 & \textbf{0.98} \\
\bottomrule
\end{tabular}
}
\end{table*}

\section{Experiments}

\subsection{Datasets and Evaluation Metrics}
\noindent\textbf{Datasets.}
We evaluate SVL on three standard shadow detection benchmarks: SBU~\cite{vicente2016large}, UCF~\cite{zhu2010learning}, and ISTD~\cite{wang2018stacked}.
SBU~\cite{vicente2016large} is a large-scale dataset with 4,089 training images and 638 test images.
It covers diverse outdoor scenes with complex illumination and shadow patterns.
We follow the official split for training and evaluation on SBU.
To assess cross-dataset generalization, we directly test the model trained on SBU on the UCF test set (110 images), which contains cluttered backgrounds and visually ambiguous shadow regions.
ISTD~\cite{wang2018stacked} contains 1,330 training images and 540 test images from 135 scenes, with diverse shadow shapes, illumination conditions, and background materials.
We follow the official split for training and evaluation.

\noindent\textbf{Evaluation Metrics.}
We report Balanced Error Rate (BER) as the primary quantitative metric.
This metric explicitly accounts for the severe class imbalance between shadow and non-shadow pixels, making it a standard evaluation metric in shadow detection.
It is defined as:
\begin{equation}
\mathrm{BER} = \left( 1 - \frac{1}{2} \left( \frac{TP}{TP+FN} + \frac{TN}{TN+FP} \right) \right) \times 100,
\end{equation}
where $TP$, $TN$, $FP$, and $FN$ are the numbers of true positives, true negatives, false positives, and false negatives, respectively.
Lower BER is better.

\subsection{Implementation Details}
\label{sec:impl}
We implement SVL in PyTorch, using a frozen DINOv3 ViT-L/16~\cite{simeoni2025dinov3} as the image encoder and a frozen CLIP text encoder~\cite{radford2021learning} for text encoding.
We extract multi-level features from blocks $\{5,11,17,23\}$ under 0--23 indexing, sampled at roughly uniform intervals to cover early, mid-level, and late representations, and additionally use block~2 as a shallow skip feature for boundary refinement.
Only lightweight modules are trained, resulting in fewer than $1\%$ trainable parameters.
Training is conducted for $10{,}000$ iterations with a batch size of $16$ using SGD~\cite{ruder2016overview} (momentum $0.9$, weight decay $5\times10^{-4}$).
The learning rate is initialized to $5\times10^{-3}$ and decayed using a polynomial schedule with power $0.9$.
Training takes approximately 40 minutes on four NVIDIA RTX 3090 GPUs.
For inference, we apply a fully connected Conditional Random Field (CRF)~\cite{krahenbuhl2011efficient} as a post-processing step to improve the spatial smoothness of the predicted shadow score maps.

\subsection{Comparison with Existing Methods}


\begin{figure*}[t]
	\centering
	\includegraphics[width=1.0\linewidth]{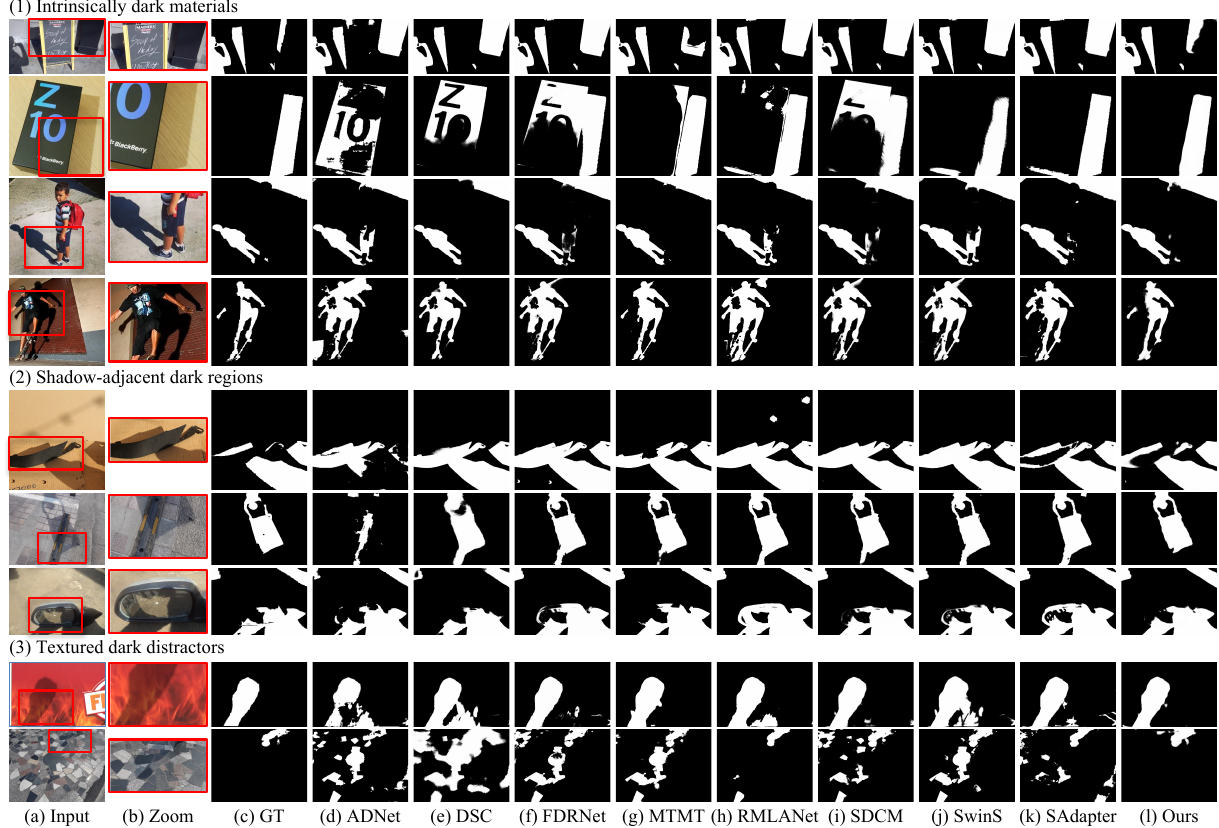}
	\caption{Qualitative comparison with representative shadow detection methods on more challenging cases with shadow--non-shadow ambiguity. From top to bottom: (1) intrinsically dark materials, (2) shadow-adjacent dark regions, and (3) textured dark distractors. SwinS and SAdapter are short for SwinShadow and ShadowAdapter, respectively.}
	\label{fig:qua2}
\end{figure*}

\noindent\textbf{Quantitative Results.}
Table~\ref{tab:quantitative} reports comparisons with existing shadow detection methods on SBU~\cite{vicente2016large}, UCF~\cite{zhu2010learning}, and ISTD~\cite{wang2018stacked}, including both early baselines and more recent methods with stronger context modeling or auxiliary semantic cues.
On SBU~\cite{vicente2016large}, SVL achieves the lowest overall BER and the lowest BER on shadow regions, while remaining competitive on non-shadow regions.
SVL also outperforms recent baselines such as SwinShadow~\cite{wang2024swinshadow} and STNet~\cite{sun2025structure}, as well as methods that incorporate auxiliary semantic cues, including SaT~\cite{zhou2024semantic} and ShadowAdapter~\cite{jie2025shadowadapter}.
These results support the hypothesis that using language as an explicit semantic reference improves discrimination when shadows and non-shadow regions exhibit substantial appearance overlap.
On UCF~\cite{zhu2010learning}, a standard benchmark for cross-dataset evaluation, SVL attains the lowest BER and the lowest shadow-region BER.
UCF contains cluttered backgrounds and visually ambiguous dark regions, where distribution shift and local appearance ambiguity make visual cues alone insufficient for shadow discrimination.
This result suggests that explicit semantic reference is beneficial when domain shift and appearance ambiguity reduce the reliability of purely visual cues.
On ISTD~\cite{wang2018stacked}, SVL achieves a competitive BER.
We observe that ISTD is collected in a relatively controlled setting and exhibits cleaner shadow boundaries, so performance on this benchmark appears to depend more on fine-grained boundary localization than on concept-level discrimination.
Under this setting, SVL remains competitive with a frozen visual backbone, and the overall results are consistent with the view that its advantage becomes more apparent when semantic ambiguity is the primary challenge.
Overall, SVL performs stably across benchmarks, with its advantages most evident in more challenging settings with visual ambiguity or dataset shift, consistent with our analysis that shadow detection can benefit from explicit semantic reference.

\begin{table*}[t]
\centering
\renewcommand{\arraystretch}{1.1}
\setlength{\tabcolsep}{6pt}
\caption{Ablation study on key components of SVL.
``BER'' denotes the balanced error rate.
``S'' and ``NS'' denote the BER values on shadow and non-shadow regions, respectively.
Best results are highlighted in \textbf{bold}.
``$\downarrow$'' indicates lower is better.}
\label{tab:ablation_components}
\scalebox{0.88}{
\begin{tabular}{l c c c ccc ccc ccc}
\toprule
\multirow{2}{*}{\textbf{Setting}} &
\multicolumn{3}{c}{\textbf{Components}} &
\multicolumn{3}{c}{\textbf{SBU}} &
\multicolumn{3}{c}{\textbf{UCF}} &
\multicolumn{3}{c}{\textbf{ISTD}} \\
\cmidrule(lr){2-4}
\cmidrule(lr){5-7} \cmidrule(lr){8-10} \cmidrule(lr){11-13}
& \textbf{Global} & \textbf{Local} & \textbf{Ratio} &
BER$\downarrow$ & S$\downarrow$ & NS$\downarrow$ &
BER$\downarrow$ & S$\downarrow$ & NS$\downarrow$ &
BER$\downarrow$ & S$\downarrow$ & NS$\downarrow$ \\
\midrule
(1) Visual baseline
&  &  & 
& 3.65 & 4.92 & 2.38
& 7.78 & 10.28 & 5.28
& 1.92 & 1.00 & 2.84 \\

(2) + Global only
& \ding{51} &  & 
& 2.94 & 3.58 & 2.29
& 6.51 & 6.39 & 6.62
& 1.61 & 1.99 & 1.22 \\

(3) + Local only
&  & \ding{51} & 
& 2.75 & 3.64 & 1.86
& 6.14 & 7.66 & \textbf{4.62}
& 1.12 & 1.04 & 1.20 \\

(4) + Global + Local
& \ding{51} & \ding{51} & 
& 2.58 & 2.65 & 2.51
& 6.03 & 6.94 & 5.12
& 1.06 & 0.94 & 1.19 \\

(5) + Scene-level ratio (w/o CRF)
& \ding{51} & \ding{51} & \ding{51}
& 2.46 & 3.10 & \textbf{1.82}
& 5.70 & 5.22 & 6.19
& 1.04 & \textbf{0.77} & 1.31 \\

(6) + Scene-level ratio (w/ CRF)
& \ding{51} & \ding{51} & \ding{51}
& \textbf{2.43} & \textbf{2.24} & 2.62 
& \textbf{5.69} & \textbf{4.72} & 6.66 
& \textbf{0.91} & 0.84 & \textbf{0.98} \\
\bottomrule
\end{tabular}
}
\end{table*}

\noindent\textbf{Qualitative Results.}
Figure~\ref{fig:qua2} presents qualitative comparisons on challenging cases with shadow--non-shadow ambiguity.
We organize these cases into three typical categories: intrinsically dark materials, shadow-adjacent dark regions, and textured dark distractors. 
In such cases, relying on visual appearance often confuses dark non-shadow regions with shadows. 
Compared with representative shadow detection methods, SVL produces more complete shadow masks while introducing fewer false positives in visually ambiguous regions.
By incorporating explicit semantic reference and cross-modal consistency at both global and local levels, SVL suppresses shadow-like responses in non-shadow areas while preserving the structure of true shadow regions.
Together with the quantitative results in Table~\ref{tab:quantitative}, these results suggest that the advantage of SVL is most apparent when explicit semantic reference is needed to distinguish true shadows from shadow-like non-shadow areas.

\subsection{Ablation Studies}
Table~\ref{tab:ablation_components} presents a component-wise ablation study on SBU~\cite{vicente2016large}, UCF~\cite{zhu2010learning}, and ISTD~\cite{wang2018stacked}.
The visual-only baseline achieves BERs of 3.65, 7.78, and 1.92 on SBU, UCF, and ISTD, respectively.
Adding only global semantic injection reduces the BER to 2.94/6.51/1.61, indicating that image-level semantic guidance provides effective semantic grounding beyond appearance cues.
Using only the local semantic constraint also yields consistent gains, reaching 2.75/6.14/1.12, suggesting that patch-level semantic guidance helps suppress shadow-like responses in non-shadow regions.
On UCF, the two semantic components exhibit complementary behavior: global semantic injection yields a lower shadow-region error, whereas the local semantic constraint achieves the lowest non-shadow-region error.
Combining them further reduces the overall BER to 2.58/6.03/1.06 on SBU/UCF/ISTD, confirming that image-level and patch-level semantic guidance are complementary.
Adding the scene-level ratio objective yields further gains, most notably on UCF, where the BER decreases from 6.03 to 5.70, highlighting the benefit of incorporating an explicit scene-level cue.
Finally, CRF~\cite{krahenbuhl2011efficient} post-processing further reduces the BER to 2.43, 5.69, and 0.91 on SBU, UCF, and ISTD, respectively, yielding the best overall results.
Overall, the ablation results show that global semantic injection, local semantic constraint, and the scene-level ratio objective provide complementary benefits, while CRF offers additional refinement at the final prediction stage.

\begin{table*}[t]
\centering
\renewcommand\arraystretch{1.08}
\setlength{\tabcolsep}{5pt}
\caption{Hard-case subset evaluation.
SBU-Hard and UCF-Hard denote the hard-case subsets constructed from SBU and UCF, respectively.
We report BER over all pixels (All: BER$\downarrow$) and within the darkest 20\% pixels of each image (D20: BER$\downarrow$, FPR$\downarrow$, Prec$\uparrow$).
Best results are highlighted in \textbf{bold}.
}
\label{tab:hardcase_all_ber_dark20}
\scalebox{0.96}{
\begin{tabular}{l c ccc c ccc c ccc}
\toprule
\multirow{2}{*}{\textbf{Method}}
& \multicolumn{4}{c}{\textbf{SBU-Hard}}
& \multicolumn{4}{c}{\textbf{UCF-Hard}}
& \multicolumn{4}{c}{\textbf{SBU-Hard+UCF-Hard}} \\
\cmidrule(lr){2-5} \cmidrule(lr){6-9} \cmidrule(lr){10-13}
& \textbf{All} & \multicolumn{3}{c}{\textbf{D20}}
& \textbf{All} & \multicolumn{3}{c}{\textbf{D20}}
& \textbf{All} & \multicolumn{3}{c}{\textbf{D20}} \\
& BER$\downarrow$ & BER$\downarrow$ & FPR$\downarrow$ & Prec$\uparrow$
& BER$\downarrow$ & BER$\downarrow$ & FPR$\downarrow$ & Prec$\uparrow$
& BER$\downarrow$ & BER$\downarrow$ & FPR$\downarrow$ & Prec$\uparrow$ \\
\midrule
ADNet~\cite{le2018a+}              & 17.04 & 19.14 & 13.69 & 57.81 & 18.31 & 23.97 & 25.94 & 26.55 & 17.19 & 19.94 & 15.48 & 52.88 \\
DSC~\cite{hu2018direction}         & 12.93 & 14.09 &  7.93 & 71.44 & 17.51 & 20.94 & 17.23 & 34.46 & 13.65 & 14.94 &  9.30 & 66.26 \\
DSD~\cite{zheng2019distraction}    &  6.16 & 11.43 & 14.39 & 61.26 & 13.30 & 17.38 & 26.39 & 29.45 &  7.27 & 12.30 & 16.15 & 56.57 \\
FDRNet~\cite{zhu2021mitigating}       &  5.73 & 11.57 & 16.84 & 58.05 & 11.22 & 18.10 & 29.76 & 27.43 &  6.56 & 12.52 & 18.73 & 53.48 \\
MTMT~\cite{chen2020multi}          &  6.95 & 10.23 &  9.13 & 70.73 & 11.73 & 14.69 & 17.53 & 37.68 &  7.70 & 10.87 & 10.36 & 66.29 \\
RMLANet~\cite{jie2023rmlanet}      &  4.92 & 10.37 & 14.63 & 61.48 & 10.13 & 13.16 & 18.75 & 37.22 &  5.74 & 10.73 & 15.23 & 58.59 \\
SDCM~\cite{zhu2022single}          &  6.70 & 10.00 &  9.20 & 70.69 &  9.36 & 13.94 & 18.48 & 37.09 &  7.09 & 10.63 & 10.56 & 66.03 \\
ShadowAdapter~\cite{jie2025shadowadapter} &  4.79 &  8.84 &  8.81 & 72.01 & 10.14 & 15.21 & 28.07 & 29.49 &  5.59 & 10.00 & 11.63 & 64.41 \\
SwinShadow~\cite{wang2024swinshadow}      &  6.04 &  9.65 &  9.57 & 70.12 &  8.29 & 15.90 & 23.82 & 31.72 &  6.33 & 10.63 & 11.66 & 64.06 \\
\midrule
Ours    & \textbf{3.88} & \textbf{6.71} & \textbf{7.27} & \textbf{76.25} &
                \textbf{7.64} & \textbf{11.29} & \textbf{11.86} & \textbf{47.50} &
                \textbf{4.46} & \textbf{7.22} & \textbf{7.94} & \textbf{73.01} \\
\bottomrule
\end{tabular}
}
\end{table*}

\subsection{Hard-Case Evaluation: Shadow vs.\ Dark Non-Shadow}
\label{sec:hardcase}
A major challenge in shadow detection lies in semantic ambiguity, as visually dark non-shadow regions may closely resemble cast shadows.
To evaluate a model's ability to resolve this ambiguity, we construct a dedicated hard-case subset and report metrics that explicitly measure the confusion between dark non-shadow regions and true shadows.

\subsubsection{Hard-case subset construction}
Given an image $\mathbf{I}_n$ with pixel domain $\Omega_n$, we denote the shadow region $S_n\subset\Omega_n$, the non-shadow region $\bar{S}_n=\Omega_n\setminus S_n$, and the brightness at location $\mathbf{x}$ by $V(\mathbf{x})\in[0,1]$ (HSV-$V$ channel).
We compute an image-adaptive darkness threshold $\tau_n^{(p)}$ as the $p$-th percentile of $V(\mathbf{x})$.
A smaller $p$ focuses on the darkest pixels.
We then quantify the proportion of dark non-shadow pixels as
\begin{equation}
r_n^{(p)} = \frac{\left|\left\{\mathbf{x}\in\bar{S}_n : V(\mathbf{x}) < \tau_n^{(p)}\right\}\right|}{|\bar{S}_n|}.
\end{equation}
Intuitively, $r_n^{(p)}$ measures the fraction of the non-shadow region whose brightness falls below $\tau_n^{(p)}$, which captures cases where dark non-shadow areas are most likely to be confused with cast shadows.
To reduce sensitivity to the percentile choice, we compute $r_n^{(p)}$ for $p\in\{5,10,15\}$ and rank images in descending order of $r_n^{(p)}$ for each $p$ (rank~1 is the hardest case).
We use the mean rank across the three percentiles as the selection criterion and construct the hard subset by selecting the 20\% images with the smallest mean rank.

\subsubsection{Hard-case results}
We evaluate all methods on the hard-case subset using BER over all pixels, together with BER, FPR, and precision within the darkest 20\% pixels of each image.
Table~\ref{tab:hardcase_all_ber_dark20} summarizes the results.
On the all-pixel evaluation, SVL achieves the lowest BER on SBU, UCF, and the merged subset, indicating stronger robustness under shadow--dark non-shadow ambiguity.
While all-pixel BER reflects overall performance, the darkest-20\% subset focuses on the most ambiguous regions, where dark non-shadow areas are most likely to be misclassified as shadow.
In this setting, FPR and precision are particularly informative for characterizing false-positive suppression.
Across SBU, UCF, and the merged subset, SVL achieves the lowest BER and FPR together with the highest precision on the darkest-20\% pixels, indicating fewer false positives in visually dark non-shadow regions.
Overall, these results suggest that SVL can suppress false positives in visually dark regions more effectively.

\subsection{Analysis of the Text Reference}
To further understand the role of language in SVL, we analyze the text reference from two perspectives, namely semantic specificity and prompt robustness.
Table~\ref{tab:text_reference_analysis} reports results on both the standard SBU and UCF test sets and the more challenging hard-case subset introduced in Section~\ref{sec:hardcase}.
Specifically, we report BER on the standard SBU and UCF test sets, and for the merged SBU-Hard+UCF-Hard subset, we report BER over all pixels together with BER, FPR, and precision within the darkest 20\% pixels of each image.

\begin{table}[t]
\centering
\scriptsize
\renewcommand{\arraystretch}{1.06}
\setlength{\tabcolsep}{3.6pt}
\caption{Text-reference analysis on standard and hard-case evaluations.
``All'' and ``D20 BER'' denote BER over all pixels and over the darkest 20\% pixels of each image, respectively.
Best results are highlighted in \textbf{bold}.}
\label{tab:text_reference_analysis}
\resizebox{\columnwidth}{!}{
\begin{tabular}{l l c c c c c c}
\toprule
\multirow{2}{*}{\textbf{ID}} & \multirow{2}{*}{\textbf{Reference}}
& \multicolumn{1}{c}{\textbf{SBU}}
& \multicolumn{1}{c}{\textbf{UCF}}
& \multicolumn{4}{c}{\textbf{SBU-Hard+UCF-Hard}} \\
\cmidrule(lr){3-3} \cmidrule(lr){4-4} \cmidrule(lr){5-8}
& & \textbf{All} & \textbf{All} & \textbf{All} & \textbf{D20 BER} & \textbf{D20 FPR} & \textbf{D20 Prec} \\
\midrule
\multicolumn{8}{l}{\textit{Semantic specificity}} \\
\midrule
T0 & Shadow prompts         & \textbf{2.43} & \textbf{5.69} & \textbf{4.46} & \textbf{7.22} & 7.94 & 73.01 \\
T1 & Dark-region desc.      & 3.11 & 7.62 & 5.48 & 9.08 & 10.84 & 66.28 \\
T2 & Illum.-nonshadow desc. & 2.88 & 6.26 & 5.97 & 8.35 & 7.66 & 72.19 \\
T3 & Irrelevant text        & 2.75 & 6.35 & 5.26 & 8.51 & 8.00 & 72.32 \\
T4 & Learnable token        & 3.04 & 6.88 & 6.38 & 9.55 & 9.87 & 67.88 \\
\midrule
\multicolumn{8}{l}{\textit{Prompt robustness}} \\
\midrule
P1 & Keyword prompts        & 2.51 & 5.77 & 4.65 & 8.38 & 9.03 & 69.13 \\
P2 & Definition prompts     & 2.53 & 6.01 & 4.82 & 8.02 & 8.90 & 70.56 \\
P3 & Attribute prompts      & 2.46 & 5.82 & 4.91 & 7.99 & \textbf{6.67} & \textbf{75.77} \\
\bottomrule
\end{tabular}
}
\end{table}

\subsubsection{Semantic Specificity of the Text Reference}
To verify whether SVL benefits from the semantic content of the text reference, we compare the default shadow-related prompts (T0) with four alternative references: generic dark-region descriptions (T1), illumination-related but non-shadow descriptions (T2), irrelevant text (T3), and a learnable token (T4), as summarized in Table~\ref{tab:text_reference_analysis}.
Overall, the default shadow-related prompts (T0) provide the strongest performance across the standard benchmarks and the hard-case evaluation.
Replacing them with generic dark-region descriptions (T1) causes the largest degradation, suggesting that these prompts provide a misleading semantic cue by reducing shadow to generic darkness.
Illumination-related but non-shadow descriptions (T2) recover part of the performance, but remain inferior to T0 because they do not correctly specify the target concept of cast shadow.
Irrelevant text (T3) is in some cases less harmful than T1, suggesting that incorrectly describing dark regions as shadows is more detrimental than using text that is simply unrelated to the task.
Replacing the text reference with a learnable token (T4) further degrades performance, indicating that an auxiliary learnable vector cannot substitute for an explicit semantically grounded reference.
Overall, SVL benefits most from text references that are well aligned with cast shadows, especially under shadow--dark non-shadow ambiguity.

\subsubsection{Prompt Robustness}
We verify whether the model predictions are sensitive to the wording of shadow-related text references.
Specifically, we evaluate three prompt variants: keyword prompts (P1), definition prompts (P2), and attribute prompts (P3).
The default prompt setting is identical to T0.
Overall, SVL exhibits limited sensitivity to prompt wording in terms of overall BER.
On SBU, the results remain within a narrow range (2.43--2.53), suggesting that the model is not tied to hand-crafted text.
On UCF, the variation is slightly larger (5.69--6.01), suggesting that the choice of prompt wording has a more visible effect under cross-dataset evaluation.
A similar trend is observed on the hard-case subset, where the all-pixel BER remains relatively stable across prompt variants, although the default prompts still achieve the best overall BER.
The darkest 20\% pixels focus on the most ambiguous dark regions.
Under this evaluation, different prompt formulations lead to a modest trade-off between false-positive suppression and overall balance.
In particular, attribute prompts (P3) achieve the lowest FPR and the highest precision, but the default prompts remain best in overall BER across both the standard and hard-case evaluations.

Overall, these two studies show that SVL benefits most from text references with the correct shadow-related semantics, while remaining robust to common wording variations when the semantic content is appropriate.



\section{Conclusion}
In this paper, we presented SVL, a parameter-efficient vision--language framework for shadow detection that incorporates language as an explicit semantic reference through global semantic injection, local semantic constraint, and scene-level shadow ratio regression.
Extensive experiments on multiple benchmarks, including dedicated hard-case evaluations, show that SVL achieves strong overall performance and improved robustness in visually ambiguous regions.
The empirical results suggest that the main benefit of SVL lies in concept-level disambiguation, particularly in reducing false positives on visually dark non-shadow regions, while the text-reference analyses further indicate that this gain depends on semantically relevant shadow descriptions rather than merely on adding a text branch.
Overall, these findings highlight the value of explicit semantic grounding for shadow detection and suggest that robust prediction requires both concept-level guidance and pixel-level detail preservation.

{\small
	\bibliographystyle{ieee_fullname}
	\bibliography{ref}

\begin{thebibliography}{10}\itemsep=-1pt

\bibitem{alayrac2022flamingo}
Jean-Baptiste Alayrac, Jeff Donahue, Pauline Luc, Antoine Miech, Iain Barr,
  Yana Hasson, Karel Lenc, Arthur Mensch, Katherine Millican, Malcolm Reynolds,
  et~al.
\newblock Flamingo: a visual language model for few-shot learning.
\newblock In {\em Proceedings of the Advances in Neural Information Processing
  Systems}, pages 23716--23736, 2022.

\bibitem{alshammari2019multi}
Naif Alshammari, Samet Ak{\c{c}}ay, and Toby~P Breckon.
\newblock Multi-task learning for automotive foggy scene understanding via
  domain adaptation to an illumination-invariant representation.
\newblock {\em arXiv preprint arXiv:1909.07697}, 2019.

\bibitem{chen2021semi}
Xiaokang Chen, Yuhui Yuan, Gang Zeng, and Jingdong Wang.
\newblock Semi-supervised semantic segmentation with cross pseudo supervision.
\newblock In {\em Proceedings of the IEEE Conference on Computer Vision and
  Pattern Recognition}, pages 2613--2622, 2021.

\bibitem{chen2023make}
Xiao-Diao Chen, Wen Wu, Wenya Yang, Hongshuai Qin, Xiantao Wu, and Xiaoyang
  Mao.
\newblock Make segment anything model perfect on shadow detection.
\newblock {\em IEEE Transactions on Geoscience and Remote Sensing},
  61(1):1--13, 2023.

\bibitem{chen2020multi}
Zhihao Chen, Lei Zhu, Liang Wan, Song Wang, Wei Feng, and Pheng-Ann Heng.
\newblock A multi-task mean teacher for semi-supervised shadow detection.
\newblock In {\em Proceedings of the IEEE Conference on Computer Vision and
  Pattern Recognition}, pages 5611--5620, 2020.

\bibitem{chi2026cross}
Kaichen Chi, Wei Jing, Junjie Li, Qiang Li, and Qi Wang.
\newblock Cross-modal spherical aggregation for weakly supervised remote
  sensing shadow removal.
\newblock {\em IEEE Transactions on Multimedia}, 28:813--824, 2026.

\bibitem{gu2021open}
Xiuye Gu, Tsung-Yi Lin, Weicheng Kuo, and Yin Cui.
\newblock Open-vocabulary object detection via vision and language knowledge
  distillation.
\newblock {\em arXiv preprint arXiv:2104.13921}, 2021.

\bibitem{guo2017efficient}
Jianting Guo, Peijia Zheng, and Jiwu Huang.
\newblock An efficient motion detection and tracking scheme for encrypted
  surveillance videos.
\newblock {\em ACM Transactions on Multimedia Computing, Communications, and
  Applications}, 13(4):1--23, 2017.

\bibitem{guo2011single}
Ruiqi Guo, Qieyun Dai, and Derek Hoiem.
\newblock Single-image shadow detection and removal using paired regions.
\newblock In {\em Proceedings of the IEEE Conference on Computer Vision and
  Pattern Recognition}, pages 2033--2040, 2011.

\bibitem{hosseinzadeh2018fast}
Sepideh Hosseinzadeh, Moein Shakeri, and Hong Zhang.
\newblock Fast shadow detection from a single image using a patched
  convolutional neural network.
\newblock In {\em Proceedings of the International Conference on Intelligent
  Robots and Systems}, pages 3124--3129, 2018.

\bibitem{hou2019large}
Le Hou, Tom{\'a}s F~Yago Vicente, Minh Hoai, and Dimitris Samaras.
\newblock Large scale shadow annotation and detection using lazy annotation and
  stacked cnns.
\newblock {\em IEEE Transactions on Pattern Analysis and Machine Intelligence},
  43(4):1337--1351, 2019.

\bibitem{hu2023cross}
Peng Hu, Zhenyu Huang, Dezhong Peng, Xu Wang, and Xi Peng.
\newblock Cross-modal retrieval with partially mismatched pairs.
\newblock {\em IEEE Transactions on Pattern Analysis and Machine Intelligence},
  45(8):9595--9610, 2023.

\bibitem{hu2021revisiting}
Xiaowei Hu, Tianyu Wang, Chi-Wing Fu, Yitong Jiang, Qiong Wang, and Pheng-Ann
  Heng.
\newblock Revisiting shadow detection: A new benchmark dataset for complex
  world.
\newblock {\em IEEE Transactions on Image Processing}, 30(1):1925--1934, 2021.

\bibitem{hu2018direction}
Xiaowei Hu, Lei Zhu, Chi-Wing Fu, Jing Qin, and Pheng-Ann Heng.
\newblock Direction-aware spatial context features for shadow detection.
\newblock In {\em Proceedings of the IEEE Conference on Computer Vision and
  Pattern Recognition}, pages 7454--7462, 2018.

\bibitem{huang2009moving}
Jia-Bin Huang and Chu-Song Chen.
\newblock Moving cast shadow detection using physics-based features.
\newblock In {\em Proceedings of the IEEE Conference on Computer Vision and
  Pattern Recognition}, pages 2310--2317, 2009.

\bibitem{inoue2020learning}
Naoto Inoue and Toshihiko Yamasaki.
\newblock Learning from synthetic shadows for shadow detection and removal.
\newblock {\em IEEE Transactions on Circuits and Systems for Video Technology},
  31(11):4187--4197, 2020.

\bibitem{jie2023rmlanet}
Leiping Jie and Hui Zhang.
\newblock Rmlanet: Random multi-level attention network for shadow detection
  and removal.
\newblock {\em IEEE Transactions on Circuits and Systems for Video Technology},
  33(12):7819--7831, 2023.

\bibitem{jie2025shadowadapter}
Leiping Jie and Hui Zhang.
\newblock Shadowadapter: Adapting segment anything model with auto-prompt for
  shadow detection.
\newblock {\em Expert Systems with Applications}, 273(1):126809, 2025.

\bibitem{jung2009efficient}
Cl{\'a}udio~Rosito Jung.
\newblock Efficient background subtraction and shadow removal for monochromatic
  video sequences.
\newblock {\em IEEE Transactions on Multimedia}, 11(3):571--577, 2009.

\bibitem{krahenbuhl2011efficient}
Philipp Kr{\"a}henb{\"u}hl and Vladlen Koltun.
\newblock Efficient inference in fully connected crfs with gaussian edge
  potentials.
\newblock In {\em Proceedings of the Advances in Neural Information Processing
  Systems}, pages 109--117, 2011.

\bibitem{le2018a+}
Hieu Le, Tomas F~Yago Vicente, Vu Nguyen, Minh Hoai, and Dimitris Samaras.
\newblock A+d net: Training a shadow detector with adversarial shadow
  attenuation.
\newblock In {\em Proceedings of the European Conference on Computer Vision},
  pages 662--678, 2018.

\bibitem{li2022blip}
Junnan Li, Dongxu Li, Caiming Xiong, and Steven Hoi.
\newblock Blip: Bootstrapping language-image pre-training for unified
  vision-language understanding and generation.
\newblock In {\em Proceedings of the International Conference on Machine
  Learning}, pages 12888--12900, 2022.

\bibitem{li2022grounded}
Liunian~Harold Li, Pengchuan Zhang, Haotian Zhang, Jianwei Yang, Chunyuan Li,
  Yiwu Zhong, Lijuan Wang, Lu Yuan, Lei Zhang, Jenq-Neng Hwang, et~al.
\newblock Grounded language-image pre-training.
\newblock In {\em Proceedings of the IEEE Conference on Computer Vision and
  Pattern Recognition}, pages 10965--10975, 2022.

\bibitem{lin2022dual}
Yijie Lin, Yuanbiao Gou, Xiaotian Liu, Jinfeng Bai, Jiancheng Lv, and Xi Peng.
\newblock Dual contrastive prediction for incomplete multi-view representation
  learning.
\newblock {\em IEEE Transactions on Pattern Analysis and Machine Intelligence},
  45(4):4447--4461, 2022.

\bibitem{liu2023visual}
Haotian Liu, Chunyuan Li, Qingyang Wu, and Yong~Jae Lee.
\newblock Visual instruction tuning.
\newblock In {\em Proceedings of the Advances in Neural Information Processing
  Systems}, pages 34892--34916, 2023.

\bibitem{liu2023decoupled}
Jiawei Liu, Qiang Wang, Huijie Fan, Wentao Li, Liangqiong Qu, and Yandong Tang.
\newblock A decoupled multi-task network for shadow removal.
\newblock {\em IEEE Transactions on Multimedia}, 25:9449--9463, 2023.

\bibitem{maxwell2008bi}
Bruce~A Maxwell, Richard~M Friedhoff, and Casey~A Smith.
\newblock A bi-illuminant dichromatic reflection model for understanding
  images.
\newblock In {\em Proceedings of the IEEE Conference on Computer Vision and
  Pattern Recognition}, pages 1--8, 2008.

\bibitem{mohajerani2019shadow}
Sorour Mohajerani and Parvaneh Saeedi.
\newblock Shadow detection in single rgb images using a context preserver
  convolutional neural network trained by multiple adversarial examples.
\newblock {\em IEEE Transactions on Image Processing}, 28(8):4117--4129, 2019.

\bibitem{mukhoti2023open}
Jishnu Mukhoti, Tsung-Yu Lin, Omid Poursaeed, Rui Wang, Ashish Shah, Philip~HS
  Torr, and Ser-Nam Lim.
\newblock Open vocabulary semantic segmentation with patch aligned contrastive
  learning.
\newblock In {\em Proceedings of the IEEE Conference on Computer Vision and
  Pattern Recognition}, pages 19413--19423, 2023.

\bibitem{niu2022boundary}
Kunpeng Niu, Yanli Liu, Enhua Wu, and Guanyu Xing.
\newblock A boundary-aware network for shadow removal.
\newblock {\em IEEE Transactions on Multimedia}, 25:6782--6793, 2022.

\bibitem{qian2024omni}
Zeheng Qian, Wen Wu, Xian-Tao Wu, and Xiao-Diao Chen.
\newblock Omni-supervised shadow detection with vision foundation model.
\newblock {\em Journal of Visual Communication and Image Representation},
  100(1):104146, 2024.

\bibitem{qiaounder}
Xiaotian Qiao, Ke Xu, Xianglong Yang, Ruijie Dong, Xiaofang Xia, and Jiangtao
  Cui.
\newblock Under the shadow: Exploiting opacity variation for fine-grained
  shadow detection.
\newblock In {\em Proceedings of the Advances in Neural Information Processing
  Systems}, pages 1--13, 2025.

\bibitem{radford2021learning}
Alec Radford, Jong~Wook Kim, Chris Hallacy, Aditya Ramesh, Gabriel Goh,
  Sandhini Agarwal, Girish Sastry, Amanda Askell, Pamela Mishkin, Jack Clark,
  et~al.
\newblock Learning transferable visual models from natural language
  supervision.
\newblock In {\em Proceedings of the International Conference on Machine
  Learning}, pages 8748--8763, 2021.

\bibitem{ruder2016overview}
Sebastian Ruder.
\newblock An overview of gradient descent optimization algorithms.
\newblock {\em arXiv preprint arXiv:1609.04747}, 2016.

\bibitem{sanin2012shadow}
Andres Sanin, Conrad Sanderson, and Brian~C Lovell.
\newblock Shadow detection: A survey and comparative evaluation of recent
  methods.
\newblock {\em Pattern Recognition}, 45(4):1684--1695, 2012.

\bibitem{simeoni2025dinov3}
Oriane Sim{\'e}oni, Huy~V Vo, Maximilian Seitzer, Federico Baldassarre, Maxime
  Oquab, Cijo Jose, Vasil Khalidov, Marc Szafraniec, Seungeun Yi, Micha{\"e}l
  Ramamonjisoa, et~al.
\newblock Dinov3.
\newblock {\em arXiv preprint arXiv:2508.10104}, 2025.

\bibitem{sun2025structure}
Wanlu Sun, Liyun Xiang, and Wei Zhao.
\newblock Structure-aware transformer for shadow detection.
\newblock {\em IET Image Processing}, 19(1):e70031, 2025.

\bibitem{tarvainen2017mean}
Antti Tarvainen and Harri Valpola.
\newblock Mean teachers are better role models: Weight-averaged consistency
  targets improve semi-supervised deep learning results.
\newblock In {\em Proceedings of the Advances in Neural Information Processing
  Systems}, pages 1195--1204, 2017.

\bibitem{tiwari2016survey}
Arti Tiwari, Pradeep~Kumar Singh, and Sobia Amin.
\newblock A survey on shadow detection and removal in images and video
  sequences.
\newblock In {\em Proceedings of the IEEE International Conference on Cloud
  Security and Big Data Engineering}, pages 518--523, 2016.

\bibitem{valanarasu2023fine}
Jeya Maria~Jose Valanarasu and Vishal~M Patel.
\newblock Fine-context shadow detection using shadow removal.
\newblock In {\em Proceedings of the IEEE Winter Conference on Applications of
  Computer Vision}, pages 1705--1714, 2023.

\bibitem{vicente2016large}
Tom{\'a}s F~Yago Vicente, Le Hou, Chen-Ping Yu, Minh Hoai, and Dimitris
  Samaras.
\newblock Large-scale training of shadow detectors with noisily-annotated
  shadow examples.
\newblock In {\em Proceedings of the European Conference on Computer Vision},
  pages 816--832, 2016.

\bibitem{wang2019moving}
Bingshu Wang, Yong Zhao, and CL~Philip Chen.
\newblock Moving cast shadows segmentation using illumination invariant
  feature.
\newblock {\em IEEE Transactions on Multimedia}, 22(9):2221--2233, 2019.

\bibitem{wang2024language}
Hongqiu Wang, Wei Wang, Haipeng Zhou, Huihui Xu, Shaozhi Wu, and Lei Zhu.
\newblock Language-driven interactive shadow detection.
\newblock In {\em Proceedings of the ACM International Conference on
  Multimedia}, pages 5527--5536, 2024.

\bibitem{wang2018stacked}
Jifeng Wang, Xiang Li, and Jian Yang.
\newblock Stacked conditional generative adversarial networks for jointly
  learning shadow detection and shadow removal.
\newblock In {\em Proceedings of the IEEE Conference on Computer Vision and
  Pattern Recognition}, pages 1788--1797, 2018.

\bibitem{wang2024swinshadow}
Yonghui Wang, Shaokai Liu, Li Li, Wengang Zhou, and Houqiang Li.
\newblock Swinshadow: Shifted window for ambiguous adjacent shadow detection.
\newblock {\em ACM Transactions on Multimedia Computing, Communications, and
  Applications}, 20(11):1--20, 2024.

\bibitem{wang2018densely}
Yupei Wang, Xin Zhao, Yin Li, Xuecai Hu, Kaiqi Huang, and NLPR CRIPAC.
\newblock Densely cascaded shadow detection network via deeply supervised
  parallel fusion.
\newblock In {\em Proceedings of the International Joint Conferences on
  Artificial Intelligence}, page~6, 2018.

\bibitem{wang2023detect}
Yonghui Wang, Wengang Zhou, Yunyao Mao, and Houqiang Li.
\newblock Detect any shadow: Segment anything for video shadow detection.
\newblock {\em IEEE Transactions on Circuits and Systems for Video Technology},
  34(5):3782--3794, 2023.

\bibitem{wu2025pay}
Xian-Tao Wu, Xiao-Diao Chen, Hongyu Chen, Wen Wu, Weiyin Ma, and Haichuan Song.
\newblock Pay more attention to dark regions for faster shadow detection.
\newblock {\em Computer Vision and Image Understanding}, 263(1):104589, 2026.

\bibitem{xu2022semi}
Haiming Xu, Lingqiao Liu, Qiuchen Bian, and Zhen Yang.
\newblock Semi-supervised semantic segmentation with prototype-based
  consistency regularization.
\newblock In {\em Proceedings of the Advances in Neural Information Processing
  Systems}, pages 26007--26020, 2022.

\bibitem{yang2024shadow}
Heejun Yang, Oh-Hyeon Choung, and Yuseok Ban.
\newblock Shadow removal for enhanced nighttime driving scene generation.
\newblock {\em Applied Sciences}, 14(23):10999, 2024.

\bibitem{yang2023silt}
Han Yang, Tianyu Wang, Xiaowei Hu, and Chi-Wing Fu.
\newblock Silt: Shadow-aware iterative label tuning for learning to detect
  shadows from noisy labels.
\newblock In {\em Proceedings of the IEEE International Conference on Computer
  Vision}, pages 12687--12698, 2023.

\bibitem{zheng2019distraction}
Quanlong Zheng, Xiaotian Qiao, Ying Cao, and Rynson~WH Lau.
\newblock Distraction-aware shadow detection.
\newblock In {\em Proceedings of the IEEE Conference on Computer Vision and
  Pattern Recognition}, pages 5167--5176, 2019.

\bibitem{zhou2024semantic}
Kai Zhou, Jing-Long Fang, Wen Wu, Yan-Li Shao, Xing-Qi Wang, and Dan Wei.
\newblock Semantic-aware transformer for shadow detection.
\newblock {\em Computer Vision and Image Understanding}, 240(1):103941, 2024.

\bibitem{zhou2023semantically}
Yiyang Zhou, Qinghai Zheng, Shunshun Bai, and Jihua Zhu.
\newblock Semantically consistent multi-view representation learning.
\newblock {\em Knowledge-Based Systems}, 278(1):110899, 2023.

\bibitem{zhu2010learning}
Jiejie Zhu, Kegan~GG Samuel, Syed~Z Masood, and Marshall~F Tappen.
\newblock Learning to recognize shadows in monochromatic natural images.
\newblock In {\em Proceedings of the IEEE Computer Society Conference on
  Computer Vision and Pattern Recognition}, pages 223--230, 2010.

\bibitem{zhu2018bidirectional}
Lei Zhu, Zijun Deng, Xiaowei Hu, Chi-Wing Fu, Xuemiao Xu, Jing Qin, and
  Pheng-Ann Heng.
\newblock Bidirectional feature pyramid network with recurrent attention
  residual modules for shadow detection.
\newblock In {\em Proceedings of the European Conference on Computer Vision},
  pages 121--136, 2018.

\bibitem{zhu2021mitigating}
Lei Zhu, Ke Xu, Zhanghan Ke, and Rynson~WH Lau.
\newblock Mitigating intensity bias in shadow detection via feature
  decomposition and reweighting.
\newblock In {\em Proceedings of the IEEE International Conference on Computer
  Vision}, pages 4702--4711, 2021.

\bibitem{zhu2022single}
Yurui Zhu, Xueyang Fu, Chengzhi Cao, Xi Wang, Qibin Sun, and Zheng-Jun Zha.
\newblock Single image shadow detection via complementary mechanism.
\newblock In {\em Proceedings of the ACM International Conference on
  Multimedia}, pages 6717--6726, 2022.

\end{thebibliography}
}

\vfill
\end{document}